\documentstyle[11pt,lingmacros]{article}
  \newlength\titlebox 
\newcommand{\bit}{\begin{itemize}}
\newcommand{\eit}{\end{itemize}} \newcommand{\ben}{\begin{enumerate}}
\newcommand{\een}{\end{enumerate}} \newcommand{\bqt}{\begin{quote}}
\newcommand{\eqt}{\end{quote}} \newcommand{\bc}{\begin{center}}
\newcommand{\ec}{\end{center}} \newcommand{\bdes}{\begin{description}}
\newcommand{\edes}{\end{description}}
\newcommand{\btab}{\begin{tabular}} \newcommand{\etab}{\end{tabular}}

\newcommand{\she}{{\bf she}}\newcommand{\her}{{\bf her}}\newcommand{\he}{{\bf he}}
\newcommand{\his}{{\bf his}}

\newcommand{\their}{{\bf their}}
\begin{document}
\input{psfig}
\title{Intrasentential Centering: A Case Study}
\author{Megumi Kameyama\\Artificial Intelligence Center and\\
The Center for the Study of Language and Information,\\SRI International\\
333 Ravenswood Ave., Menlo Park, CA 94025, U.S.A.\\ megumi\verb+@+ai.sri.com}
\date{
\begin{scriptsize}
1997 (in press). In Walker, M., A. Joshi, and
E. Prince, eds., {\it Centering Theory in Discourse}, Oxford
University Press, Oxford.
\end{scriptsize}
}
\maketitle

\section{Introduction}

One of the necessary extensions to the centering model is a mechanism
to handle pronouns with intrasentential antecedents.  Existing
centering models deal only with discourses consisting of simple
sentences. It leaves unclear how to delimit center-updating utterance
units and how to process complex utterances consisting of multiple
clauses. In this paper, I will explore the extent to which a
straightforward extension of an existing intersentential centering
model contributes to this effect. I will motivate an approach that
breaks a complex sentence into a hierarchy of center-updating units
and proposes the preferred interpretation of a pronoun in its local
context arbitrarily deep in the given sentence structure.  This approach
will be substantiated with examples from naturally
occurring written discourses.

\section{Centering Model}\label{model}

Centering (Grosz, Joshi, and Weinstein, 1983, 1986, 1995) synthesizes
two lines of research. One is the {\it discourse centering model}
(Joshi and Kuhn, 1979; Joshi and Weinstein, 1981) concerned with the
issue of grammar-based control of discourse inference.  The basic
claim is that discourse has a {\it monadic tendency} --- the tendency
to be `about' one thing at a time, which is indicated in utterance
structures. The other is the {\it discourse focusing model} (Grosz,
1977; Sidner, 1979; 1983; Grosz and Sidner, 1986) concerned with the
discourse structure and meaning associated with the discourse
participants' intentions, attentions, and linguistic choices.

Centering is part of an overall theory of discourse structure and
meaning (Grosz and Sidner, 1986) that distinguishes among three
components of discourse structure --- a linguistic structure, an
intentional structure, and an attentional state --- and two levels of
discourse coherence --- global and local.  {\it Attentional state}
models the discourse participants' focus of attention determined by
the intentional and linguistic structures at any given point in the
discourse. It has global and local components corresponding to the two
levels of discourse coherence. Centering models the local-level
component of attentional state  --- how the speaker's linguistic
choices for representing propositional contents affect the {\em
inference load placed upon the hearer} in discourse processing.

In centering, an utterance in discourse (not a sentence in isolation)
has entities called {\em centers} that link the utterance with other
utterances in the discourse segment. They are a set of {\em
forward-looking centers} (Cf) partially ordered by relative
prominence.  One member of the Cf may be the {\em backward-looking
center} (Cb) that connects with a member of the Cf of the previous
utterance.  The speaker's linguistic choices define centering
transitions that affect the local coherence of the discourse.  In
English discourse, pronouns and grammatical subjects are the main
indicators of centering transitions (Grosz et al., 1983, 1986;
Kameyama, 1985, 1986; Brennan, Friedman, and Pollard, 1987).  {\em
Unstressed} pronouns are primarily used to indicate the Cb in
English-type languages, whereas it is the zero pronominals in
Japanese-type languages according to the syntactic typology of
pronominal forms (Kameyama, 1985: Ch.1).

\subsection{The Basic Centering Model}

The core insight of the centering model fits well into a general
dynamic discourse processing architecture (Kameyama, 1992a, 1992b,
1996).\footnote{The dynamic discourse processing architecture 
here is in line with dynamic semantic theories such as 
by Kamp (1981), Heim (1982),
and Groenendijk and Stokhof (1991).}  
In the following reformulation of centering within this
architecture, I will make a crucial use of {\it salience} as a
primitive notion underlying {\it centering dynamics}:\footnote{Note
that Constraint-3 and Rule-2 of the centering theory in the
introduction of this volume are not assumed in this
reformulation. These specifics are discussed in the next subsection.}

\bit
\item {\it Discourse} is a sequence of utterances $U_1,...,U_n$
produced by one or more discourse participants.
\item Each utterance $U_i$ defines a transition between an {\it input
context} $C_{i-1}$ and an {\it output context} $C_i$, where a context
$C_k$ is a dynamically evolving cognitive information state (more or
less) shared by the discourse participants.
\item A component in context $C_k$ is the {\it attentional
state} $A_k$ consisting of a set of currently ``open'' propositions
with the associated entities.\footnote{The ``open'' propositions and
entities in $A_k$ are those into which a new utterance content can potentially
be integrated.  $A_k$ corresponds to the {\it focus space stack}
(Grosz and Sidner, 1986) whose internal structure is determined by the
current position in the overall discourse structure.} 
The entities in $A$ are partially ordered by {\it
salience} (see below).
\item The most salient subpart of $A_k$ is the centering state 
$CEN_k$ consisting of a set of propositions with associated entities
called the {\it forward-looking centers} $Cf_k$. 
\item One
of the entities in $Cf_k$ may be the {\it backward-looking center}
$Cb_k$, or {\it the Center}, the central entity that the discourse is
currently about. $Cb_k$ is also a member of $Cf_{k-1}$, but is not necessarily
the same as $Cb_{k-1}$.
\item With centering rules and constraints, 
each utterance $U_i$ defines a transition between an input 
centering state $CEN_{i-1}$ and an output centering state $CEN_i$
controlling the inferences involved in updating the context with each
new utterance.  
\eit

I will henceforth distinguish two kinds of input context $C_{i-1}$ for
an arbitrary utterance $U_i$. It is an {\it intersentential context}
when $U_{i-1}$ and $U_i$ belong to different ``sentences'' in the standard
syntactic sense. It is an {\it intrasentential context} when $U_{i-1}$
and $U_i$ belong to the same sentence.

A discourse describes situations, eventualities, and entities,
together with the relations among them. The attentional state $A$
represents a dynamically updated snapshot of their {\it salience}. I
assume the property {\it salient} to be a primitive representing the
{\it partial order} among a set of entities in $A$.  Salience is
gradient and relative.  A certain absolute degree of salience may not
be achieved by any entities in a given $A$, but there is always a set
of {\it maximally salient} entities, which we can call the ``Cm'' for
convenience. The Cm is often, but not necessarily, a
singleton set.  Note that the Cm differs from the existing centering
notion of the Cb (or Cp), although they are related to one another. I
argue that the property of the Cm, over and above that of the Cb (and
Cp), elucidates the centering dynamics, as discussed in the remainder
of this section.

Various factors affect salience dynamics
--- including utterance forms, discourse participants' purposes and
perspectives, and the perceptually salient objects in the utterance
situation. The specification of salience dynamics is a crucial step
toward a formal theory of discourse pragmatics, and centering focuses
on the interrelations between the center dynamics and utterance forms.

Centering stipulates that the entities in the Cf are generally more
salient than other entities in $A$, and if one of these entities is
the Center (Cb), it is the most salient entity. We state this
stipulation as a defeasible preference:\footnote{Defeasibility is
indicated by ``normally''.}
\bdes
\item[CENTER] The Center is normally more salient than other entities in
the same attentional state.
\edes 

\subsection{Linguistic Correlates of Salience}

We can now state a set of linguistic correlates to salience dynamics
in terms of additional defeasible preferences (Kameyama, 1996).

Two default linguistic hierarchies are crucial in centering dynamics
--- the {\it grammatical function hierarchy} (GF ORDER) and the {\it
nominal expression type hierarchy} (EXP ORDER).\footnote{Both linguistic hierarchies are
in fact recurrent in functional and typological studies of language.
The GF ORDER closely resembles Keenan and Comrie's (1977)
Accessibility Hierarchy, Givon's (1979) Topicality Hierarchy, and
Kuno's (1987) Thematic Hierarchy, all of which predict the preferred
syntactic structure for describing the things that a sentence is
``mainly about'' within and across languages.  The EXP ORDER resembles
the linguistic correlates of Gundel, Hedberg, and Zacharski's (1993)
Givenness Hierarchy, which is closely related to Prince's (1981)
Familiarity Scale, which predicts the relative {\it degrees} of
accessibility of referents. It is of interest that virtually the same
hierarchies are relevant to the computational interest in how grammar
controls inferences in language use.}  
\bdes
\item[GF ORDER:] Given a hierarchy [{\sc SUBJECT $>$ OBJECT $>$ OBJECT2 $>$
Others}], an entity realized by a higher-ranked phrase is normally
more salient in the {\it output} attentional state.
\item[EXP ORDER:] Given a hierarchy [{\sc Zero
Pronominal $>$ Pronoun $>$ Definite NP $>$ Indefinite
NP}],\footnote{There is a pragmatic difference between stressed and
unstressed pronouns, which should be accounted for by an independent
treatment of stress --- for example, in terms of a preference reversal
function (Kameyama, 1994). This paper concerns only unstressed
pronouns.}  an entity realized by a higher-ranked expression type is
normally more salient in the {\it input} attentional state.
\edes

For each utterance $U_i$, EXP ORDER and GF ORDER predict the relative
salience of entities in the {\it input} and {\it output} attentional
states $A_{i-1}$ and $A_i$, respectively. 
GF ORDER in $U_i$ directly affects 
the {\it output} attentional state $A_i$.  It corresponds to the major
determinant of the Cf ordering in centering.  EXP ORDER in $U_i$
is an assumption or presupposition about the {\it
input} attentional state $A_{i-1}$. It generalizes the centering
Rule-1.
EXP ORDER also predicts the relative salience of entities in the
{\it output} attentional state $A_i$ since these assumed salience
levels are also often {\it accommodated} into the context (see Lewis,
1979). 

Another centering stipulation is that at most one entity is given the
Center status at any one attentional state. I propose the following
defeasible preference as a specialization of EXP ORDER:%
\bdes
\item[EXP CENTER:] An expression of the highest-ranked type in EXP
ORDER normally realizes the Center in the output attentional
state.\footnote{When an utterance contains more than one
highest-ranked expression types, the Center is realized by either (1) the
Cb-chaining one or (2) the one with the higher-ranked GF, in this
preference order.}
\edes
EXP CENTER departs from Constraint-3 in defining 
what makes an entity a Cb. Constraint-3 relies on the Cf ordering,
whereas EXP CENTER relies on EXP ORDER. The following example
illustrates the difference: 
\eenumsentence{\label{party}
\item[1.] John went to Jim's party. 
\item[2.] {\it he} was very pleased to see John again. 
\item[3.] {\it he} had just recovered from a stressful week at work.
} After (\ref{party})-2, EXP CENTER makes Jim the Cb, whereas
Constraint-3 makes John the Cb. The preferred referent of {\it he} in
(\ref{party})-3 is then Jim under the present approach due to GF
ORDER and EXP CENTER whereas it is John under the centering algorithm
due to the preference for CONTINUE over SMOOTH-SHIFT.

EXP CENTER also enables two choices in what expression types are
associated with the Center status.  The ``highest-ranked type'' in
EXP CENTER can be interpreted as either relative to each utterance or
absolute in all utterances. Under the relative interpretation, a
nonpronominal expression type can also output the Center as long as
there are no pronouns in the same utterance.  Under the absolute
interpretation, only the pronominals (either zero or overt, depending
on the syntactic type of the language) can output the Center.  I will
take the absolute interpretation in this paper following Kameyama
(1985, 1986), based on the rationale that the choice of the
highest-ranked pronominal forms in a language should reflect a
certain absolute sense of salience threshold.

This paper will focus on English centering.  Since matrix subjects and
objects cannot be omitted in the English-type
language,\footnote{Except in a telegraphic register.} the
highest-ranked expression type that outputs the Center is the
(unstressed) pronoun. From EXP ORDER, it follows that a pronoun {\it
normally} realizes a {\it maximally salient entity} in the input
attentional state.  Since it is a defeasible preference, a pronoun may
also realize a submaximally salient entity under certain conditions.
Known such conditions fall under three classes:
\ben
\item Another overriding preference gives a
different interpretation of the pronoun. Kameyama (1996), in
motivating general preference classes with overriding relationships,
points out that specific commonsense causal knowledge can generally override
salience-based local coherence preferences. Garden-path effects of
centering studied by Hudson D'Zmura (1988) show that this overriding
can require a considerable effort when 
the salience-based preference is particularly ``strong''.

\item The
denotational range imposed by the pronoun's grammatical features rule
out the maximally salient entity. For instance, {\it he} would not
resolve to a maximally salient female person entity.

\item Sortal constraints on the pronoun's argument position coming
from the verb rule out the maximally salient entity. For instance, in
``I learned {\it it}'', the referent of {\it it} is sortally
inconsistent with, say, a salient dog entity.  \een

\subsection{Centering Preference Interactions}

One of the notable features of the present reformulation of centering
is that the transition types per se play no role in determining the
preferred interepretation of utterances. They are mere ``labels'' on
centering transitions that result from the interactions of the
centering preferences stated in CENTER, GF ORDER, EXP
ORDER, and EXP CENTER. For instance, we can classify transitions into
``establishing'' and ``chaining'' (Kameyama, 1985,1986) as
follows.\footnote{What I have previously called {\it retain}
(Kameyama, 1985, 1986, 1988) is now called {\it chain}. } The
Center is ``established'' when a pronoun picks a salient non-Center
in the input context and makes it the Center in the output context.
It corresponds to both types of SHIFT in centering.
It is ``chained'' when a pronoun picks the Center in the input context
and makes it the Center in the output context. It corresponds to both
CONTINUE and RETAIN in centering. The preference for CONTINUE over
RETAIN is a consequence of the general preference for {\it
determinate} maximal preference as discussed below.

\subsubsection*{(In)Determinate Maximal Salience}

The interaction of CENTER, GF ORDER, EXP ORDER, and EXP CENTER
accounts for canonical intersentential centering examples. A novel
aspect of the present setup is that each centering state contains the
{\it set} of {\it maximally salient entities}
(Cm), and the Cm's (in)determinacy predicts the
corresponding (in)determinacy of the preferred interpretation of a
pronoun.\footnote{Discussions with Becky Passonneau helped clarify
this perspective.} The Cm is determinate when it is a singleton set,
and indeterminate when it is a set of two or more entities. The
interpretation preference of a pronoun is determinate when it
converges on a single maximal preference, and indeterminate when its
maximal preference is a set of equally preferred entities.\footnote{Hoffman
(this volume) discusses the indeterminacy of RETAIN transitions.}
  EXP
ORDER states that a pronoun's preferred interpretation is the Cm, so
the (in)determinacy in the Cm predicts the (in)determinacy in a
pronoun's preferred interpretation (unless other overriding factors
remove the indeterminacy).

This is how it works. There are two independent sources for the Cm,
the highest-ranked GF defined in GF ORDER and the Cb.  When they converge
on the same entity, it is the single member of the Cm, but when they
diverge on two different entities, both are in the Cm. In English
centering, for instance, a subject Center achieves a convergence, and
a nonsubject Center leads to a divergence in most cases.\footnote{Nonsubjects 
can be made more salient than subjects in special syntactic constructions 
such as clefting and topicalization.}
This is illustrated below.
\eenumsentence{\label{babar1}
\item[1.] Babar went to a bakery.
\\\hspace{3cm}$A_1$:[[Babar$>$Bakery]$_{Cf}$\ldots]

\item[2.] $he$ greeted the baker.
\hspace{1.5cm}$he$ := Babar
\\\hspace{3cm}$A_2$:[[[Babar]$^{Subj}_{Center}$$>$Baker]$_{Cf}$$>$Bakery\ldots]

\item[3.] $he$ pointed at a blueberry pie.
\hspace{1.5cm}$he$ := Babar$\prec$Baker
} Example (\ref{babar1}) shows the effects of determinate salience
ranking ($>$) in terms of a chain of subject Centers. 
The preferred value of the pronoun {\it he} in (\ref{babar1})-3 is
also determinate (indicated by determinate preference ordering
$\prec$) with a strong preference ($\chi^2_{df=1}=13$, $p<.001$) 
for 13 native English speakers. The next example shows the effects of indeterminate salience ranking:
\eenumsentence{\label{babar2}
\item[1.] Babar went to a bakery.
\\\hspace{3cm}$A_1$:[[Babar$>$Bakery]$_{Cf}$\ldots]

\item[2.] The baker greeted $him$.
\hspace{1.5cm}$him$ := Babar
\\\hspace{3cm}$A_2$:[[Baker$<>$[Babar]$^{Obj}_{Center}$]$_{Cf}$$>$Bakery\ldots]

\item[3.] $he$ pointed at a blueberry pie.
\hspace{1.5cm}$he$ := Baker$\prec_?$Babar
} 
In (\ref{babar2}), the
salience ranking in $A_2$ is indeterminate ($<>$) and the preferred
value of {\it he} shifts to the baker, although this preference is weak
($\chi^2_{df=1}=3.77$, $.05<p<.10$) for 13 nonoverlapping speakers.

The notion of (in)determinate attentional preference is not entirely
new to centering, but it has not been given appropriate
recognition.  If we call the highest-ranked GF in the GF ORDER the
``preferred Center'' (Cp) as in Brennan et al. (1987), we are talking
about the convergence (Cb=Cp) and divergence (Cb$\neq$Cp) of the Cb
and Cp in the {\it input} attentional state here.  In the centering
algorithm (Brennan et al., 1987; Walker, Iida, and Cote, 1994), the same
convergence-divergence distinction takes on quite a different role.
First of all, the distinction is in the {\it output} state of
utterance interpretation, separating out CONTINUE and RETAIN on
one hand, and SMOOTH-SHIFT and ROUGH-SHIFT on the other, and a fixed
preference ordering among these transition states predicts the
preferred interpretation of a pronoun. Second of all, the preferred
pronoun interpretation is always determinate. The algorithm is defined
so that no indeterminacy arises.\footnote{The only indeterminacy in
their centering algorithm comes from ``optional'' rules such as the
Zero Topic Assignment rule in Walker et al. (1994). It is unclear what
the logical/cognitive status of these optional rules are. The best
characterization is that their application is random.} 

\subsubsection*{Parallelism Preference}

The weak preference in (\ref{babar2})-3 comes from the interaction
of centering preference and the separate preference for structural
parallelism. In Kameyama (1996), it was hypothesized that when a
determinate attentional preference and structural parallelism
preference conflict, the former overrides the latter, but
when the attentional preference is indeterminate, the parallelism
preference kicks in to give rise to a weak parallelism preference.
This parallelism preference is loosely stated in the following defeasible preference:
\bdes
\item[PARA:] Two adjacent utterances in discourse seek maximal 
parallelism.\footnote{For a specific
definition of parallelism, see, e.g., Pruest (1992), Kameyama (1986).}
\edes

Integrating both monadic and parallelism effects
is crucial for making accurate predictions about local
coherence preferences based on utterance structures. 
The property-sharing constraint on centering (Kameyama, 1985,
1986) was a proposal for this integration.\footnote{See Passonneau (1993)
for empirical supports for the property-sharing constraint.}
Here, the integration is achieved at a higher level of preference interactions.

Parallelism effects on local discourse coherence have been
neglected in centering because centering focuses on orthogonal
``monadic'' effects.\footnote{See Suri and McCoy (1994) for a similar criticism.} 
For example,
the centering algorithm incorrectly predicts a definite preference for {\it he}
:= Babar in (\ref{babar2})-3 because CONTINUE (Cb=Cp=Babar) is
preferred over SMOOTH-SHIFT (Cb=Cp=Baker) after (\ref{babar2})-2
(Cb=Babar, Cp=Baker). This example contradicts the claim made in
support of the centering algorithm that ``structural parallelism is a
consequence of our ordering the Cf list by grammatical functions and
the preference for continuing over retaining'' (Brennan et al., 1987,
p.157). 

The remainder of the paper addresses the issue of how to extend the
existing intersentential centering model to discourses consisting of
arbitrarily complex sentences.

\section{Issues of Intrasentential Centering}

The existing centering models say nothing explicit about how to
analyze complex sentences.\footnote{Walker (1989) discusses the need
for this extension to centering, and suggests several constraints.}
What are relative preferences for a pronoun
to realize entities in the intersentential or intrasentential context?
What counts as evidence?  Does a centering model contribute to
elucidating an aspect of intrasentential binding phenomena overlooked
in purely syntactic approaches? In the rest of the paper, I will
propose a centering model that makes an explicit claim about complex
sentences, motivating it from theoretical and empirical grounds.
All the example discourses in the remainder of the paper come from the
Brown corpus (Francis and Kucera, 1982).

To substantiate the importance of intrasentential (pronominal) anaphora, I have
compared the counts of intrasentential and intersentential anaphora
with 3rd person pronouns in nineteen randomly selected seventeen-sentence
discourses in the Brown corpus that contain numerous {\it he}-type
pronouns.\footnote{So-called pleonastic {\it it} was excluded.} Among
 255 3rd person pronouns in total, of which 184 (65.8\%) are {\it
he}-type pronouns, 149 (58.4\%) have their antecedents in the same
sentence, 100 (39.2\%) have their antecedents in the immediately preceding
sentence, and 6 (2.4\%) have their antecedents in the second most recent sentence. All the antecedents precede the pronouns.
There are thus more intrasentential than intersentential
anaphoric dependencies in these naturally occurring written
texts.\footnote{When a sentence has a set of coreferring pronouns, I
have counted at most one of them to be intersentential anaphora. For
instance, in ``{\it He said he was lucky,}'' there are one
intersentential and one intrasentential anaphoric dependencies.} I
will now motivate a set of extensions to centering to handle complex
sentences, and discuss a number of theoretical and computational
issues.

\subsection{Sentence-based Centering?}

There is an approach to intrasentential centering that I am rejecting.
It is to process a whole sentence at once, deciding for each pronoun
whether it realizes an entity in the intersentential discourse context
or something evoked within that sentence. I will henceforth call this
approach the {\it sentence-based centering}.  This approach first
appears attractive because it would imply that a single input
centering state functions as a control factor for a whole complex
sentence. If this is indeed the case, it would not only give a further
motivation to centering but also support the special status of a
sentence as an atomic and autonomous unit that organizes a discourse.
The following example, in fact, appears to support this
sentence-based centering analysis:\footnote{In subsequent examples,
third person nonpleonastic pronouns appear in boldface, and each
utterance ($U_k$) is labeled with a centering transition type
discussed above --- CHAIN (Cb$_{k-1}$=Cb$_k\neq$NULL), ESTABLISH
(Cb$_{k-1}\neq$Cb$_k\neq$NULL), and NULL (Cb$_k$=NULL).}

\enumsentence{\label{sutherland}
{\bf Example: Sutherland}
\ben
\item[1.] CHAIN(Cb=Sutherland): {\bf Her} entrance in Scene 2 Act 1 brought some
disconcerting applause even before {\bf she} had sung a note.

\item[2.] CHAIN(Cb=Sutherland): Thereafter the audience waxed applause happy 
but discriminating operagoers reserved judgment as {\bf her} singing
showed signs of strain, {\bf her} musicianship some questionable
procedure, and {\bf her} acting uncomfortable stylization.

\item[3.] CHAIN(Cb=Sutherland): As {\bf she} gained composure during the second act 
{\bf her} technical resourcefulness emerged stronger though {\bf she}
had already revealed a trill almost unprecedented in years of
performances of Lucia.
\een
}

Discourse example (\ex{0}) above consists of three complex sentences,
each of which contains multiple {\it she}-type pronouns all referring to
the same person, Sutherland. We can see these sentences to be all
`centrally about' Sutherland, and this ``Cb'' in the input $CEN$
determines the uniform referent for all the pronouns in each sentence.

A sentence can contain multiple pronouns with different referents,
however, and when these pronouns are all in the same type, their
disambiguation is a nightmare. An example follows.

\enumsentence{\label{sarah}
{\bf Example: Sarah}
\ben
\item[1.] ESTABLISH(Cb=Sarah): And in all likelihood by now there was more than one person in the house who knew the terms of \her\  marriage contract. 

\item[2.] CHAIN(Cb=Sarah): There was no point either in telling herself again what a fool \she\  had been. 

\item[3.] CHAIN(Cb=Sarah): {\bf She} went downstairs and received another curious shock, for when Glendora flapped into the dining room in \her\  homemade moccasins, Sarah asked \her\  when \she\  had brought coffee to \her\  room, and Glendora said \she\  hadn't. 
\een
}

The sentence of interest is (\ref{sarah})-3 repeated below.
\enumsentence{\label{glendora}
input: [$Cb=Sarah_1$]\\ $She_1$ went downstairs and received another curious
shock, for when $Glendora_2$ flapped into the dining room in $her_?$
homemade moccasins, $Sarah_1$ asked $her_2$ when $she_?$ had brought
coffee to $her_?$ room, and $Glendora_2$ said $she_?$ hadn't.  }

Here, assuming that only the two most salient female entities, Sarah and
Glendora, are involved, syntactic constraints on pronominal
coreference and the explicit names in the sentence can help determine
only 2 out of 6 pronoun references. Remaining ambiguities make $2^4
=16$ possible combinations for the sentence as a whole.  Even if
ellipsis resolution forces the two $she_?$ subjects to corefer, there
are $2^3=8$ possible combinations.  It would be desirable if the
complex sentence can be processed piece by piece, treating one or two
pronouns at a time, reducing inferential spaces and combinatorics.

\subsection{Computational Motivation}

One might argue that sentence-based centering is not very
complex because sentence grammar problems are more manageable than
discourse grammar problems. However, attachment ambiguities in
sentence grammar lead to exponential combinatorics, and processing long
and complex sentences is computationally intensive. Moreover, 
discourse structure analyses (e.g., Polanyi, 1988; Hobbs, 1990) have shown
that sentences in discourse do not form a flat structure but rather 
a tree-like structure, where sentences start
at widely different depths. In short, we know
{\it a priori} that sentence grammars are no better at handling
intrasentential anaphora. 

The approach to intrasentential centering that I will advocate here is
to break a complex sentence into a structured sequence of
subsentential units each of which is the ``utterance'' in
intersentential centering.  This approach is desirable from a
computational ground. If a complex sentence can be processed piece by
piece, then the computational load involved in utterance processing
becomes more manageable. This occurs because the utterance processing
complexity is predicted to increase exponentially with the number of
references in the utterance that must be resolved with respect to the
discourse context. This desire for {\it resource boundedness} to
curtail the computational load is the original motivation of the
centering model (Joshi and Weinstein, 1981).

\subsection{Linguistic Motivation}


As discussed in Section \ref{model}, there are parallel and monadic
tendencies that govern intersentential local discourse
coherence. Grammaticization of these tendencies in connecting clauses
within a sentence would be the prime linguistic motivation for
clause-based centering. Structural parallelism is known to govern the wellformedness of a
number of ellipsis phenomena such as gapping and verb phrase ellipsis. Control
phenomena involving unexpressed subjects in nonfinite (i.e., tenseless) clauses can be
seen as an example of grammaticized monadicity. Syntactic
``movements'' observed in topicalization, left- and
right-dislocation, and clefting also give prominence to a single entity, thus illustrating the monadic tendency. These tendencies should then also affect the preferential
aspects of clause-to-clause transitions. We expect, however, stronger
semantic, rhetorical, or causal constraints coming from explicit
clausal connectives for conjunctions and subordinations in intrasentential processing because 
explicit event relations and rhetorical relations between
clauses constrain inferences more strongly than implicit relations
``between lines'' in sentence-to-sentence discourse processing.

There is evidence that intrasentential anaphoric dependencies of
pronominals in complement clauses are also controlled by an analogical
interpertation of intersentential centering. The analogy here is
between a linear sequence of utterances $U_1,...,U_n$ and
recursive embeddings of clauses (Kameyama,
1988).
\begin{figure}
\centerline{\psfig{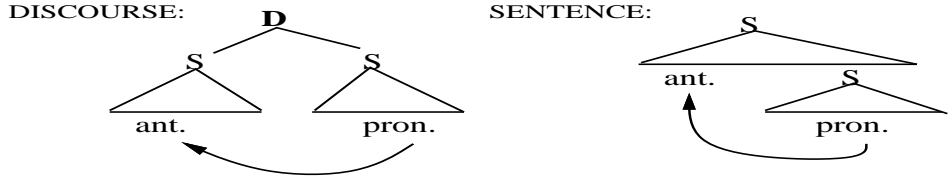}}
\caption{\label{analogy-fig}Discourse-Sentence Analogy}
\end{figure}
The preferred antecedent-pronoun pair shares a certain grammatical
property such as SUBJECT and IDENT (grammaticized speaker's point of
view in Japanese, commonly called ``Empathy'').  From this
perspective, given GF ORDER and PARA, in {\it John told Bill that he ...},
the preferred antecedent of {\it he} is {\it John}. These general
preferences, however, can be overridden by commonsense inferences. For
example, in {\it John asked Bill when he ...}, the commonsense
preference leads to {\it Bill} as the preferred antecedent of {\it
he}.


\subsection{Intrasentential Centering Hypothesis}

I will motivate the following top-level hypothesis:\footnote{Of 
interest is the fact that something like this hypothesis has been
taken for granted in the focusing approach (Grosz, 1977; Sidner, 1979;
1983; Grosz and Sidner, 1986). For instance, the following sequence of
``utterances'' is from Sidner (1983: D26):
\ben
\item Wilbur is a fine scientist and a thoughtful guy.
\item He gave me a book a while back which I really liked.
\item It was on relativity theory,
\item and talks mostly about quarks.
\item They are hard to imagine,
\item because they indicate the need for elementary field theories of
a complex nature.
\item ...
\een 
Note that 3--4 and 5--6 break up sentences into subsentential clauses
each of which updates the local focus. No definition has been given
anywhere in the literature, however, for what subsentential units
count as independent utterances. }

\enumsentence{\label{ICH}
{\bf Intrasentential Centering Hypothesis (ICH):} A complex sentence is
broken up into a set of center-updating units corresponding
to the ``utterances'' in intersentential centering.
}

\begin{figure}
\centerline{\psfig{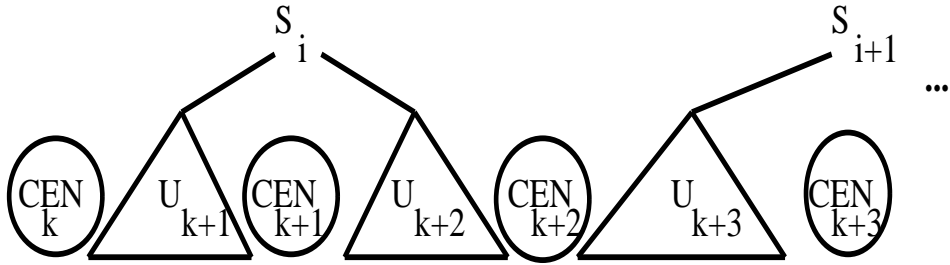}}
\caption{\label{sequence-fig}Sequential Intrasentential Centering}
\end{figure}

This hypothesis still leaves open what structure the subsentential
utterances form. It could be a flat sequence, a tree, or something
more complex. If it is a flat sequence, there is always a
single centering state, and the output center of a complex sentence is
the output of the last subsentential unit. This {\it sequential intrasentential
centering} is illustrated in Figure \ref{sequence-fig}.

\begin{figure}
\centerline{\psfig{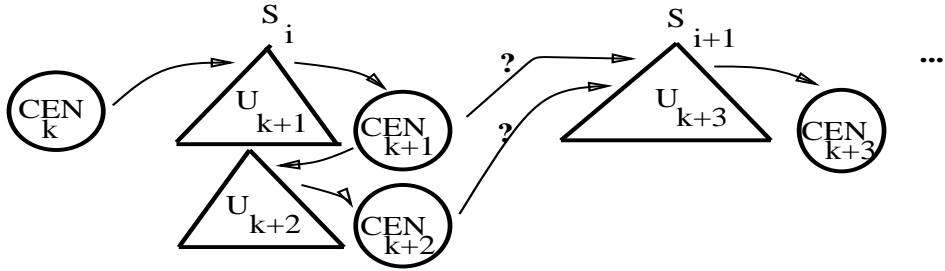}}
\caption{\label{hierarchy-fig}Hierarchical Intrasentential Centering}
\end{figure}

If it is a tree, we need to allow multiple centering states
simultaneously active at different depths of embedding. This {\it
hierarchical intrasentential centering} is illustrated in Figure
\ref{hierarchy-fig}.  The set of simultaneously active $CEN$s here lie
on the ``right open edge'' of the evolving discourse structure. At
this point, it is an open question which of the multiple
input centering states is the most prominent.

In general, the ICH in (\ref{ICH}) predicts the following:

\enumsentence{\label{ICHprediction}
{\bf Prediction from ICH:} After a complex sentence S, the subsequent
discourse favors the input center to be the one that results from
breaking up S into a (structured) sequence of subsentential units
rather than the one that results from S as a whole. In other words,
the output center of a complex sentence is the output of the last
(possibly nested)
subsentential unit in the linear ordering rather than the output
heavily influenced by the matrix clause of S alone.  } 

The validity of this hypothesis depends on how a complex sentence is
broken up into utterance units and what structure these utterance
units form.  A number of dimensions need to be considered for a full
account: nominal versus clausal units, matrix versus subordinate
clauses, the linear ordering of clauses, tensed versus untensed
clauses, conjuncts versus adjuncts, adjuncts versus complements,
direct versus indirect reports, and restrictive versus nonrestrictive
relative clauses. For each dimension, we would like to know when and
how the center is updated.

A central question here is the relation between the {\it syntactic
structure} of a complex sentence and its {\it discourse structure} in
terms of centering units. Do the centering units in a sentence form a
flat sequence or a recursively embedded hierarchical structure? If
they are hierarchical, are they isomorphic to the syntactic
hierarchy?\footnote{These questions lead to a connection with the
Linguistic Discourse Model (LDM) (Polanyi, 1988; Scha and Polanyi,
1988). In LDM, the elementary information unit in discourse is a {\it
discourse constituent unit} (dcu) that mostly corresponds to a tensed
or untensed clause. A discourse is a left-to-right growth of a
complex tree-like structure with recursive embeddings of segments, each
of which consists of a set of dcu's. Sentences correspond to subtrees
that may start from (or ``attach to'') any level of embedding in the
discourse tree.  Of interest here is to what extent the LDM converges
with an extended centering model that does both inter- and
intrasentential updating.  More specifically, is every dcu a
center-updaing unit, or vice versa?}

In sum, the following questions are central in intrasentential
centering:
\ben
\item What subsentential units update the center? 
\item Are there some embedded updates ``hidden'' from the top-level centering?
\item How are the sentence structure and discourse structure related?
\een

In the next section, I will state a number of specific hypotheses as partial answers to these questions, and motivate them with naturally occurring discourses.

\section{Clause-based Intrasentential Centering}

As a direct extension of intersentential centering, tensed clausal
units are the best place to start carving out the mechanism of
intrasentential centering, for the following reasons: 
\ben

\item Untensed
clauses are more grammatically integrated with superordinate clauses,
leaving relatively less room for pragmatics (e.g., grammatically
controlled unexpressed subjects in infinitives and gerunds).

\item Attentional state updating with tense and aspect as proposed by
Kameyama, Passonneau, and Poesio (1993) can be unified with
centering state updating with tensed clauses.
\een

I have then obtained the distribution of the antecedents from the
perspective of this tensed clause-based centering. Here, rather than
the antecedent's sentence location, we look at its tensed clause
location.  Henceforth, an utterance ($U$) is a tensed clause.
Among the 255 third person pronouns in nineteen seventeen-sentence discourses
above, (a) 83 (32.5\%) have their antecedents in the immediately
preceding utterance in the immediately preceding sentence, (b) 82
(32.2\%) have their antecedents in the immediately preceding or
superordinate utterance in the same sentence, and (c) 65 (25.5\%) have
their antecedents in the same utterance. These 230 {\it local}
dependencies account for overwhelming 90.2\% of all the pronouns in
these corpora.  Clause-based centering proposed here will make
predictions about the (b) cases that existing centering has nothing to
say about, and will refine the predictions about the (a) cases by
focusing on the clause rather than entire sentence that immediately
precedes the clause containing the pronoun in question. 45(69.0\%) of the (c) cases
are {\it possessive} pronouns that occur in tenseless clauses
or nominal expressions as in {\it its rejection}, {\it his estimate},
and {\it their revision}.\footnote{The relative salience between the
possessor and possessed still needs investigation.}
I propose that possessive pronouns are
subject to an even tighter locality constraint than nonpossessive
pronouns, and that their preferred antecedents lie {\it nearer in
their left} within their utterance units 
regardless of their grammatical functions.\footnote{Hobbs's
(1978) syntactic algorithm posits such a tight locality preference for all pronouns.}

In this section, I will propose an initial classification of clausal
relation types in terms of their association with either sequential
or hierarchical intrasentential centering. Subsection \ref{sequential}
discusses sequential centering, and Subsection \ref{hierarchical} 
discusses hierarchical centering.

\subsection{Sequential Intrasentential Centering}\label{sequential}

Tensed clauses of interest here are conjuncts and adjuncts. The
reductionist view that I take is that {\it tensed conjuncts and 
adjuncts define sequential centering structures}.  More specific
hypotheses are stated with supporting examples below.\footnote{
In the following centering analysis, I will 
annotate conjunction and subordination connectives between adjacent (sets of) 
utterances in italics. These connectives, such as {\it and, but, so, when}, 
and {\it because}, control utterance interpretation inferences by inducing 
rhetorical and discourse coherence relations. When no overt connectives appear,
there is an implicit ``and'' between utterances. The interaction between 
these connectives and attentional dynamics is an area for future research.}

\subsubsection*{Conjuncts}

The central hypothesis for tensed conjuncts is this.

\enumsentence{\label{tconj}
{\bf Tensed Conjunct Hypothesis (TConj):}
Tensed clausal conjuncts $Cl_1,...,Cl_n$ break up into a
sequence of utterances $U_1,...,U_n$ at the level of embedding at which
$Cl_1$ starts out in the given discourse structure.
TConj Example 1:
\ben
\item CHAIN: {\bf Her} mother was a Greer
\item CHAIN: {\it and} {\bf her} father's family came from the Orkney Isles.
\een

TConj Example 2:
\ben
\item NULL: Happy but discriminating operagoers reserved judgment
\item ESTABLISH: {\it as} {\bf her} singing showed signs of strain,
\item CHAIN: {\bf her} musicianship 0 some questionable procedure,
\item CHAIN: {\it and} {\bf her} acting 0 uncomfortable stylization. 
\een
}

Hypothesis TConj in (\ref{tconj}) 
predicts that after an embedded conjunct segment
closes, the input center to the subsequent discourse is the output of
the last embedded unit rather than the output of the whole sentence.
In the following example, Sutherland is established and chained as
the Cb within an adjunct segment, and she is not mentioned at all in
the top-level clauses.  The next sentence simply chains Sutherland as
the Cb, showing the natural flow from the more recent embedded segment
than from the more distant matrix clauses.

\enumsentence[(\ref{sutherland}$'$)]{
\begin{enumerate}
\item CHAIN(Cb=Sutherland): {\bf Her} entrance in Scene 2 Act 1 brought some
disconcerting applause 
\item CHAIN: {\it even before} {\bf she} had sung a note. 
\item NULL: {\it Thereafter} the audience waxed applause happy 
\item NULL: {\it but} discriminating operagoers reserved judgment 
\item ESTABLISH(Cb=Sutherland): {\it as} {\bf her} singing showed signs of strain 
\item CHAIN: {\bf her} musicianship some questionable procedure 
\item CHAIN: {\it and} {\bf her} acting uncomfortable stylization.
\item CHAIN: {\it As} {\bf she} gained composure during the second act 
\item CHAIN: {\bf her} technical resourcefulness emerged stronger 
\item CHAIN: {\it though} {\bf she} had already revealed a trill almost unprecedented in years of performances of Lucia 
\end{enumerate}
}

Under the present approach, tenseless clauses do not count as utterances in 
centering, which is restated below. The example contains nested infinitive
clauses with grammatically controlled unexpressed subjects. This is a case of 
grammaticized monadic tendency, and leaves little ambiguity.

\enumsentence{\label{tlessconj}
{\bf Tenseless Conjunct Hypothesis (TlessConj):}
Tenseless subordinate clausal conjuncts do not update the
center, and belong to the same
utterance unit as the immediately superordinate clause.
\ben
\item CHAIN: I wanted [to grab {\bf her} by the arm and beg {\bf her} [to wait,
to consider, to know for certain]]
\een
}

Syntactic conjunction demonstrates, on the other hand, grammaticized 
parallelism tendency, where parallel constituents can be elided from positions
that must be overt in a standalone sentence. 

\enumsentence{\label{cpara}
{\bf Conjunct Parallelism Hypothesis (CPara):}
Two adjacent conjuncts (tensed or tenseless) induce parallelism.
\ben
\item CHAIN: {\bf She} had held to the letter of {\bf her} contract
\item CHAIN: {\it and} 0 didn't come onto the stage. (zero-subject)
\een
(See also the second example for TConj above for an elided verb {\it showed} in
conjuncts.)
}

\subsubsection*{Adjuncts}

The central hypothesis about tensed adjunct clauses is this.

\enumsentence{\label{tadj}
{\bf Tensed Adjunct Hypothesis (TAdj):}
Tensed clausal adjuncts are utterance units separate from and
at the same level of embedding as their immediately superordinate clauses.\\
TAdj Example 1:
\ben
\item CHAIN: {\it Although} {\bf she}'s still a teenager who looks like a baby,
\item CHAIN: {\bf she} is getting married. 
\een

TAdj Example 2:
\ben
\item CHAIN: {\bf Her} entrance in Scene 2 Act 1 brought some disconcerting applause
\item CHAIN: {\it even before} {\bf she} had sung a note. 
\een
}

Tenseless adjunct clauses and phrases are analogous to tenseless conjuncts:

\enumsentence{\label{tlessadj}
{\bf Tenseless Adjunct Hypothesis (TlessAdj):}
Tenseless clausal and phrasal adjuncts belong to the same
utterance unit as the immediately superordinate clause.
\ben
\item CHAIN: [In the fullness of {\bf her} vocal splendor], {\it however},
{\bf she} could sing the famous scene magnificently.
\een
}

Hypothesis TAdj in (\ref{tadj}) 
gives a natural account of an example such as the
following where the matrix clause either establishes or chains the Cb
introduced in the preceding adjunct clause:

\enumsentence{
{\bf Example: Pearson}
\begin{enumerate}
\item NULL: {\it Although} Pearson disbelieved almost everything Lizzie said
\item ESTABLISH(Cb=Pearson): {\it and} {\bf 0} read a sinister purpose into almost
everything {\bf she} did 
\item CHAIN: {\bf he} happily accepted {\bf her} statement about Bridget as the whole truth. 
\item CHAIN: {\bf He} felt nothing further need be said about the servant girl. 
\end{enumerate}
}

This pattern of anaphoric dependency has led syntactic theories to 
``raise'' preposed adjuncts to a position dominating main clauses.
The left-to-right center
updating dynamics inside a sentence captures this dominance with precedence.

TAdj also predicts that in the case of apparent backward anaphora
such as in {\it When he woke up, Bill was very tired}, the pronoun
in the adjunct clause actually realizes an entity already in the
context, rather than anaphorically dependent on {\it Bill}. The
examples of backward anaphora I found are exactly such cases. In the following,
the pronoun in question is in 4:

\enumsentence{
{\bf Example: Kern}
\ben
\item CHAIN(Cb=Jim Kern): {\bf He} was particularly struck by a course on
Communist brainwashing.
\item NULL: Kern began reading a lot about the history and philosophy
of Communism
\item ESTABLISH(Cb=Jim Kern): {\it but} {\bf 0} never felt there was anything {\bf he} as
an individual could do about {\bf it}.
\item CHAIN(Cb=Jim Kern): {\it When} {\bf he} attended the Christian Anti Communist
Crusade school here about six months ago
\item NULL: Jim became convinced that an individual can do something
constructive in the ideological battle
\item ESTABLISH(Cb=Jim Kern): {\it and} {\bf 0} set out to do {\bf it}.
\een
}

\subsection{Hierarchical Intrasentential Centering}\label{hierarchical}

When an ``embedded discourse segment'' is created, an embedded
centering sequence $CEN_{k+1},...,CEN_{k+m}$ continues from the
current center $CEN_k$, and when it closes, the next utterance has
multiple alternative input centers. The question of a considerable
theoretical import here is the (relative) accessibility 
of the multiple levels of centering for subsequent discourse. Are they
all open? Does one of them close at any point? (See Figure
\ref{hierarchy-fig}.)  I found two significantly different types of
embedded centering in this connection. One is the reported direct
speech, and the other is the nonreport complement.

\subsubsection*{Reported Speech}

The following hypothesis seems reasonble for a number of examples:

\enumsentence{\label{speech}
{\bf Reported Speech Complement Hypothesis (Speech):}
Reported speech is an embedded centering segment that is inaccessible
to the superordinate centering level.
}

A canonical example follows. It is a sequence of reported speeches, where
each reported complement (in a syntactic sense) can be a multisentence
discourse:

\enumsentence{\label{hughes}
{\bf Example: Hughes}
\ben
\item Hughes said Monday, ``It is the apparent intention of the
Republican Party to campaign on the carcass of what {\bf they} call
Eisenhower Republicanism but the heart stopped beating and the
lifeblood congealed after Eisenhower retired.  Now {\bf he}'s gone the
Republican Party is not going to be able to sell the tattered remains
to the people of the state.''
\item Sunday {\bf he} had added, ``We can love Eisenhower the man even
if we considered {\bf him} a mediocre president but there is nothing
left of the Republican Party without {\bf his} leadership.'' 
\item Mitchell said the statement should become a major issue in the
primary and the fall campaign.
\een
}

A clause-based centering analysis of (\ref{hughes}) follows:

\enumsentence[(\ref{hughes}$'$)]{
\ben
\item NULL: Hughes said Monday, 
\ben
\item ESTABLISH(Cb=RepParty): ``It is the apparent intention of the
Republican Party to campaign on the carcass of what {\bf they} call
Eisenhower Republicanism 
\item CHAIN(Cb=RepParty): {\it but} {\bf the heart}\footnote{The 
unexpressed
possessor of this relational noun is the Republican Party. The same goes
for {\it the lifeblood} in (c).}
 stopped beating 
\item CHAIN(Cb=RepParty): {\it and} {\bf the lifeblood} congealed 
\item NULL: {\it after} Eisenhower retired.  
\item ESTABLISH(Cb=Eisenhower): {\it Now} {\bf he}'s gone 
\item NULL: the Republican Party is not going to be able to sell the tattered remains
to the people of the state.''
\een
\item ESTABLISH(Cb=Hughes): Sunday {\bf he} had added, 
\ben
\item NULL: ``We can love Eisenhower the man 
\item ESTABLISH(Cb=Eisenhower): {\it even if} we considered {\bf him} a mediocre president 
\item CHAIN(Cb=Eisenhower): {\it but} there is nothing left of the Republican Party without {\bf his} leadership.''
\een 
\item NULL: Mitchell said 
\ben
\item NULL: the statement should become a major issue in the
primary and the fall campaign.
\een
\een
}
If (\ref{hughes}) is analyzed as a
left-right flat sequence of clauses,  the
unambiguity of \he\ in the first utterance of 2 would not be predicted. 
With the clear
marking of the reported segment given with quotations, discourse {\it
popping} is unambiguous at both utterance 2 and utterance 3, and the
 salient $he$-entity within the
reported segment does not remain salient at the higher level. The
salience is `inaccessible' at the higher level, blocking the
left-to-right flow of salience dynamics.

\subsubsection*{Nonreport Complements}

The next embedded centering type is nonreport complements. My
hypotheses here are split into tensed and tenseless nonreport
complements. I hypothesize that the tensed complement gives rise to an
embedded segment whose salience ordering is accessible to the higher-level centering.  

\enumsentence{\label{comp}
{\bf Clausal Complement Hypothesis (Comp):}
Tensed clausal nonreport complements create embedded discourse segments.
\ben
\item CHAIN: {\bf Her} choice of one color means
\ben
\item CHAIN:  {\bf she} is simply enjoying the motor act of coloring 
without having reached the point of selecting suitable colors for
different objects.
\een\een
}

There are not enough examples to determine the
finer-grained preference ordering between entities realized in the embedded centering unit and those realized in the higher centering unit, however. 
The same situation holds for relative clauses. Relative clauses create embedded centering units that carry on nested centering transitions, but 
when they close, their salient entities may be accessible but
may not be salient at the higher level.
These are areas for future investigations.

On the other hand, I hypothesize that the tenseless complement does
not give rise to an embedded centering level. It is processed with the
higher clause. For instance,

\enumsentence{\label{tlesscomp}
{\bf Tenseless Complement Hypothesis (TlessComp):}
Tenseless clausal complements belong to the same utterance units as their
superordinate clauses.
\ben
\item ESTABLISH: We watched {\bf them} [set out up the hill hand in hand on a rainy day 
in {\bf their} yellow raincoats [0 to finger paint at the grammar school]].
\een
}

\section{Conclusion}

I have posited a set of plausible mechanisms to treat complex
sentences as pieces of discourse where intersentential centering
applies on subsentential `utterance' units.  
These initial hypotheses have been
motivated with naturally occurring discourse examples. Both monadic and 
parallel tendencies observed in intersentential centering have been 
found in clause-based intrasentential centering, which is a promising sign
that sentence-based and clause-based centering analyses would converge
on a uniform set of attentional notions and predictions. 

With the approach developed here, the example of ambiguous pronouns in 
(\ref{sarah})-3 is analyzed as follows:
\enumsentence[(\ref{glendora}$'$)]{
input: [Cb=Sarah$_1$]
\ben
\item CHAIN(Cb=Cp=Sarah): {\bf She}$_1$ went downstairs and received another curious
shock, 
\item ESTABLISH(Cb=Cp=Glendora): {\it for when} Glendora$_2$ flapped into the dining room in {\bf her}$_2$ homemade moccasins, 
\item CHAIN(Cb=Glendora, Cp=Sarah): Sarah$_1$ asked {\bf her}$_2$ 
\ben
\item CHAIN(Cb=Cp=Glendora): when {\bf she}$_2$ had brought coffee to {\bf her}$_1$ room, 
\een
\item NULL(Cb=NULL, Cp=Glendora): {\it and} Glendora$_2$ said 
\ben
\item ESTABLISH(Cb=Cp=Glendora): {\bf she}$_2$ hadn't.  
\een
\een
}

The main contribution of the present intrasentential centering approach is to 
reduce the amount of inferences required for utterance interpretation by making
utterances smaller. 
The tight locality preference of possessive pronouns proposed at the
beginning of Section 4 correctly predicts the reference of {\it her}
in utterance 2, but fails for {\it her} in utterance 3(a). The latter
is a case where the commonsense preference overrides the attentional
preference, presumably based on an earlier mention of coffee having
been brought to Sarah's room. In fact, utterance 3 outputs an
indeterminate centering state, where Sarah and Glendora are both
maximally salient (see Section 2.3), so the salience-based
interpretation of 3(a) is largely open for a commonsense
influence. Note, for instance, that if 3(a) were "when she should
bring coffee to her room," the local commonsense preference would
shift to opposite readings of the pronouns. These commonsense
applications are more controlled in smaller utterance units proposed
in this paper.

I hope these intrasentential centering proposals make the centering
theory more complete, and serve to be the basis for further extensions
and integrations.

\section*{References}

\noindent
\begingroup
\parindent=-2em \advance\leftskip by2em
\parskip=5pt

Brennan, Susan, Lyn Friedman, and Carl Pollard. 1987.  A Centering
Approach to Pronouns. In {\it Proceedings of the 25th Annual Meeting
of the Association for Computational Linguistics}, 155--162.


Francis, W\@. and H\@. Kucera. 1982.
Frequency Analysis of {E}nglish Usage: Lexicon and Grammar,
Houghton Mifflin, Boston, MA.

Givon, Talmy. 1979. {\it On Understanding Grammar}, Academic Press,
New York, NY.

Groenendijk, J. and M. Stokhof. 1991. Dynamic Predicate Logic. {\it
Linguistics and Philosophy} 14, 39-100.

Grosz, Barbara. 1977. The Representation and Use of Focus in Dialogue
Understanding. Technical Report 151, SRI International, 333 Ravenswood
Ave, Menlo Park, CA 94025.

Grosz, Barbara, Aravind Joshi, and Scott Weinstein. 1983.  Providing a
Unified Account of Definite Noun Phrases in Discourse.  In {\it
Proceedings of the 21st Meeting of the Association of Computational
Linguistics}, Cambridge, MA, 44--50.

Grosz, Barbara, Aravind Joshi, and Scott Weinstein. 1986.  Towards a
Computational Theory of Discourse Interpretation.  Unpublished
manuscript. [The final version appeared as Grosz et al. 1995]

Grosz, Barbara, Aravind Joshi, and Scott Weinstein. 1995. 
Centering: A Framework for Modelling
the Local Coherence of Discourse. In {\it Computational
Linguistics}, 21(2), 203--226.

Grosz, Barbara and Candy Sidner. 1986. Attention, Intention, and
the Structure of Discourse. {\it Computational Linguistics}, 12(3), 175-204. 

Gundel, Jeanette, Nancy Hedberg, and Ron Zacharski. 1993. Cognitive
Status and the Form of Referring Expressions in Discourse. {\it
Language}, 69(2), 274--307.

Heim, Irene. 1982. {\it The Semantics of Definite and Indefinite Noun
Phrases,} Ph.D. Thesis, University of Massachusetts, Amherst.


Hobbs, Jerry. 1978. Resolving Pronoun References. {\it Lingua}, 44,
311--338. Also in B.~Grosz, K.~Sparck-Jones, and B.~Webber, eds., {\it
Readings in Natural Language Processing}, Morgan Kaufmann, Los Altos,
CA, 1986, 339--352.

Hobbs, Jerry. 1990. {\it Literature and Cognition}, CSLI Lecture Note
21, CSLI, Stanford, CA.

Hoffman, Beryl. This volume. Word Order, Information Structure, and
Centering in Turkish.

Hudson D'Zmura, Susan. 1988. {\it The Structure of Discourse and
Anaphor Resolution: The Discourse Center and the Roles of Nouns and
Pronouns.} Ph.D. Thesis, University of Rochester.

Joshi, Aravind, and Steve Kuhn. 1979. Centered Logic: The Role of Entity
Centered Sentence Representation in Natural Language Inferencing. In
{\it Proceedings of International Joint Conference on Artificial
Intelligence}, Tokyo, Japan, 435--439.

Joshi, Aravind, and Scott Weinstein. 1981. Control of Inference: Role of
Some Aspects of Discoruse Structure --- Centering. In {\it Proceedings
of International Joint Conference on Artificial Intelligence},
Vancouver, Canada, 385--387.

Kameyama, Megumi. 1985. {\it Zero Anaphora: The Case of Japanese.}
Ph.D. Thesis, Stanford University.

Kameyama, Megumi. 1986. A Property-sharing Constraints in Centering.
In {\it Proceedings of the 24th Annual Meeting of
the Association for Computational Linguistics}, New York, NY, 200--206.

Kameyama, Megumi. 1988. Zero Pronominal Binding: Where Syntax and
Discourse Meet. In Poser, William, ed., {\it Papers from the Second
International Workshop on Japanese Syntax}, CSLI, Stanford, CA, 47--74.

Kameyama, Megumi. 1992a. Discourse Understanding and World Knowledge.
{\it Journal of Information Processing 15(3)}, Tokyo: Information
Processing Society of Japan, 377--385.

Kameyama, Megumi. 1992b. The Linguistic Information in Dynamic
Discourse,  Report No. CSLI--92--174, CSLI, Stanford, CA.

Kameyama, Megumi. 1994. Stressed and Unstressed Pronouns:
Complementary Preferences. In P.~Bosch and R. van der Sandt,
eds., {\it Focus and Natural Language Processing}, Institute for Logic
and Linguistics, IBM, Heidelberg, 475--484. (The final revised version to appear in 
P.~Bosch and R. van der Sandt, eds., {\it Focus: Linguistics, Cognitive, and Computational Perspectives}, Cambridge University Press, Cambridge, in press.)

Kameyama, Megumi. 1996. Indefeasible Semantics and Defeasible
Pragmatics. In Kanazawa, Makoto, Christopher Pi\~{n}on, and
Henri\"{e}tte de Swart, Eds., {\it Quantifiers, Deduction, and
Context},  CSLI, Stanford, CA, 111--138.

Kameyama, Megumi, Rebecca Passonneau. and Massimo Poesio.  1993. Temporal
Centering. In {\it Proceedings of the 31st Annual Meeting of the
Association for Computational Linguistics}, Columbus, OH, 70--77.

Kamp, Hans. 1981. A Theory of Truth and Semantic Representation. In
J. Groenendijk, T. Janssen, and M. Stokhof, eds., {\it Formal Methods in
the Study of Language}, Mathematical Center, Amsterdam, 277--322. 

Keenan, Edward, and Bernard Comrie. 1977. Noun Phrase Accessibility
and Universal Grammar. {\it Linguistic Inquiry}, 8(1), 63--100.

Kuno, Susumu. 1987. {\it Functional Syntax}, Chicago University Press, Chicago.

Lewis, David. 1979. Scorekeeping in a Language Game. {\it Journal of
Philosophical Logic}, 8, 339--359.


Passonneau, Rebecca. 1993. Getting and Keeping the Center of Attention.
In R. Weischedel and M. Bates, eds., {\it Challenges in Natural
Language Processing}, Cambridge University Press,  Cambridge. 

Polanyi, Livia. 1988. A Formal Model of Discourse Structure. {\it Journal
of Pragmatics}, 12, 601--638.

Prince, Ellen. 1981. Toward a Taxonomy of Given-New Information. In
{\it Radical Pragmatics}, Academic Press, 223--255.

Pruest, Hub. 1992. On Discourse Structuring, VP Anaphora and Gapping. PhD
Dissertation, University of Amsterdam.


Scha, Remko, and Livia Polanyi. 1988. An Augmented Context Free
Grammar of Discourse. In {\it the Proceedings of the 12th International
Conference on Computational Linguistics}, Budapest, Hungary, 22--27.


Sidner, Candace. 1979. {\it Towards a Computational Theory of Definite
Anaphora Comprehension in English Discourse.} Ph.D. Thesis (Technical
Report 537), Artificial Intelligence Laboratory, Massachusetts
Institute of Technology, Cambridge, MA.

Sidner, Candace. 1983.  Focusing in the Comprehension of Definite
Anaphora.  In M.~Brady and R.~C. Berwick, eds., {\em Computational
Models of Discourse}, The MIT Press, Cambridge, MA, 267--330.

Suri, Linda, and Kathleen McCoy. 1994. RAFT/RAPR and Centering: a
Comparison and Discussion of Problems Related to Processing Complex
Sentences. {\it Computational Linguistics}, 20(2), 301--317.

Walker, Marilyn. 1989. Evaluating Discourse Processing Algorithms. In
{\it Proceedings of the 27th Meeting of the Association of
Computational Linguistics}, Vancouver, British Columbia, Canada,
251--261.

Walker, Marilyn, Masayo Iida, and Sharon Cote. 1994. Japanese
Discourse and the Process of Centering. {\it Computational
Linguistics}, 20(2), 193--233.


\endgroup
\end{document}